\documentclass{article}

\PassOptionsToPackage{numbers, compress}{natbib}
\usepackage[nonatbib, preprint]{neurips_2022}

\usepackage[utf8]{inputenc} % allow utf-8 input
\usepackage[T1]{fontenc}    % use 8-bit T1 fonts
\usepackage[hidelinks]{hyperref}       % hyperlinks
\usepackage{url}            % simple URL typesetting
\usepackage{booktabs}       % professional-quality tables
\usepackage{amsfonts}       % blackboard math symbols
\usepackage{nicefrac}       % compact symbols for 1/2, etc.
\usepackage{microtype}      % microtypography
\usepackage{xcolor}         % colors

\usepackage{graphicx}
\usepackage{comment}
\usepackage{amsmath,amssymb} % define this before the line numbering.
\usepackage{color}
\usepackage{bm}
\usepackage{amsfonts}
\usepackage{array}
\usepackage{algorithm}
\usepackage{algorithmic}
\usepackage{booktabs}
\usepackage{float}
\usepackage{multirow}
\usepackage{mathrsfs}
\usepackage{amsfonts,amssymb} 
\usepackage{lipsum}
\usepackage{wrapfig}
\usepackage{makecell}
\usepackage{soul}

\usepackage[backend=bibtex,natbib=true,style=ieee,mincitenames=1,maxcitenames=2,maxbibnames=20,url=false,doi=false]{biblatex}
\AtEveryBibitem{
    \ifentrytype{inproceedings}{
         \clearlist{address}
         \clearlist{publisher}
         \clearname{editor}
         \clearlist{organization}
         \clearfield{url}  
         \clearfield{doi}  
         \clearfield{pages}  
         \clearlist{location}
         \clearlist{month}
     }{}
     \ifentrytype{article}{
         \clearlist{address}
         \clearlist{publisher}
         \clearname{editor}
         \clearlist{organization}
         \clearfield{url}  
         \clearfield{doi}  
         \clearlist{location}
     }{}
 }
\title{EvolveHypergraph: Group-Aware Dynamic \\Relational Reasoning for Trajectory Prediction}

\author{\textbf{Jiachen Li$^{1,}$}\thanks{indicates equal contribution} \quad \textbf{Chuanbo Hua}$^{2,*}$ \quad \textbf{Jinkyoo Park}$^{1,2}$ \quad \textbf{Hengbo Ma}$^3$ \\ \textbf{Victoria Dax}$^1$ \quad  \textbf{Mykel J. Kochenderfer}$^1$ \\
     \\
     \normalsize $^1$Stanford University \quad $^2$Korea Advanced Institute of Science and Technology\\
     $^3$University of California, Berkeley\\
     {\tt\small \{jiachen\_li, vmdax, mykel\}@stanford.edu \quad \{cbhua, jinkyoo.park\}@kaist.ac.kr} \\ {\tt\small hengbo\_ma@berkeley.edu} 
}

\begin{document}

\maketitle

\begin{abstract}
While the modeling of pair-wise relations has been widely studied in multi-agent interacting systems, its ability to capture higher-level and larger-scale group-wise activities is limited.
In this paper, we propose a group-aware relational reasoning approach (named EvolveHypergraph) with explicit inference of the underlying dynamically evolving relational structures, and we demonstrate its effectiveness for multi-agent trajectory prediction.
In addition to the edges between a pair of nodes (i.e., agents), we propose to infer hyperedges that adaptively connect multiple nodes to enable group-aware relational reasoning in an unsupervised manner without fixing the number of hyperedges.
The proposed approach infers the dynamically evolving relation graphs and hypergraphs over time to capture the evolution of relations, which are used by the trajectory predictor to obtain future states.
Moreover, we propose to regularize the smoothness of the relation evolution and the sparsity of the inferred graphs or hypergraphs, which effectively improves training stability and enhances the explainability of inferred relations.
The proposed approach is validated on both synthetic crowd simulations and multiple real-world benchmark datasets. 
Our approach infers explainable, reasonable group-aware relations and achieves state-of-the-art performance in long-term prediction.
\end{abstract}

\vspace{-0.2cm}
\section{Introduction}
\begin{figure}[!tbp]
	\centering
	\includegraphics[width=0.85\textwidth]{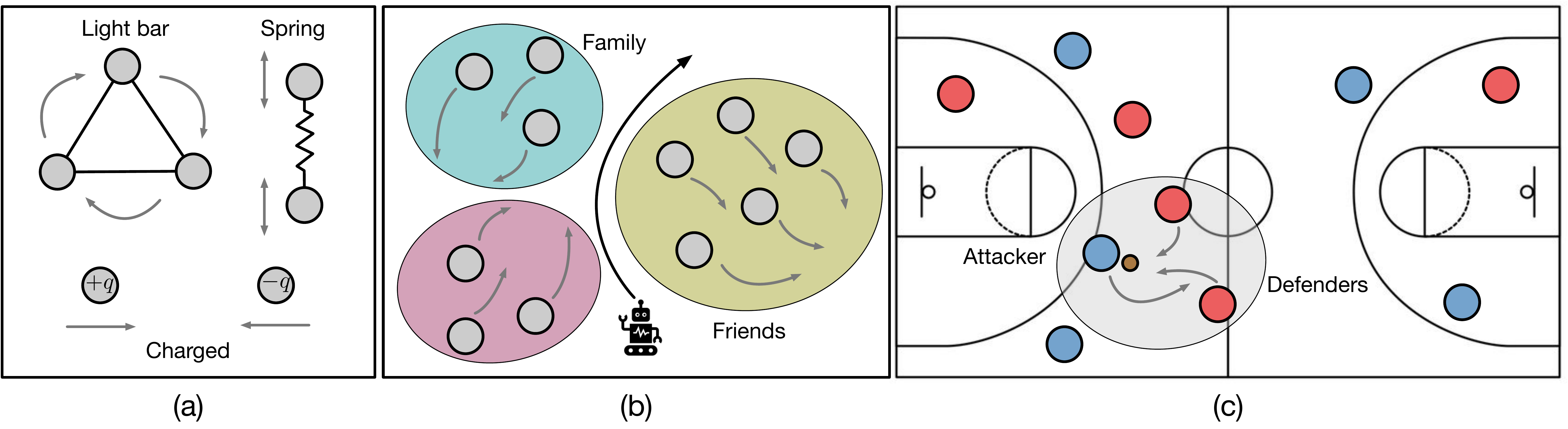}
	\vspace{-0.3cm}
	\caption{A diagram of three typical multi-agent interacting systems. The arrows indicate the agents' motions and ellipses denote different groups of interacting agents that exhibit group-wise relations. (a) Mixed particle systems with pair-wise relations. (b) Human crowds with pair-wise and group-wise relations where agents in the same group tend to have similar motion patterns. (c) Sports players with pair-wise and group-wise relations where agents in the same group may have highly distinct motion patterns. Instead of relations between particles in the form of additive forces in Figure \ref{fig:teaser}(a), the group-wise interactions between humans in Figure \ref{fig:teaser}(b)(c) are highly complicated which may not be simply modeled by pair-wise interactions. Therefore, it is necessary to model relations at both the individual level and group level explicitly. Best viewed in color.}
	\vspace{-0.4cm}
	\label{fig:teaser}
\end{figure}

\vspace{-0.1cm}
Modeling the dynamics of multi-agent interacting systems is a challenging yet significant task in various domains such as dynamics modeling for physical scene understanding \citep{battaglia2016interaction,watters2017visual,yi2019clevrer}, trajectory prediction for human-robot interaction and social robot navigation \citep{rudenko2020human,kothari2021human}, and motion forecasting for strategy planning of sports players \citep{kipf2018neural,li2020evolvegraph,graber2020dynamic}.
Examples of such systems are shown in Figure \ref{fig:teaser}.

Many research efforts in multi-agent trajectory prediction have been devoted to pair-wise relational reasoning \citep{kipf2018neural,huang2019stgat,li2020evolvegraph,graber2020dynamic,salzmann2020trajectron++,li2021rain} based on the graph representations where nodes and edges represent agents and pair-wise relations, respectively. However, the group-wise relational reasoning, which is critical for group-level interaction modeling, remains largely underexplored.
Although group activity recognition in images or videos has been widely studied in the computer vision community in recent years \citep{wu2021comprehensive,li2021groupformer,yuan2021spatio,perez2022skeleton}, these approaches heavily leverage the visual appearance information and human skeleton motions which provide informative cues for identifying group-level behaviors.
In contrast, only the historical state information is available in trajectory prediction, which makes group recognition much more difficult.
Group-LSTM \citep{bisagno2018group} uses the coherent filtering technique \citep{zhou2012coherent} to cluster the agents based on their collective movements. This group clustering strategy works well for the scenario in Figure \ref{fig:teaser}(b) where the agents in the same group behave similarly, but it may not generalize well to the scenario in Figure \ref{fig:teaser}(c) where agents in the same group can have highly distinct or even adversarial behaviors. Moreover, Group-LSTM models interactions implicitly by learning a feature embedding for each agent through a social pooling layer, which is fixed in the whole prediction horizon and difficult to explain.
To address these issues, it is necessary to design a dynamic, explainable, and flexible mechanism for group-aware relational reasoning.

A natural representation for group-wise relations is a hypergraph where each hyperedge can connect more than two nodes or agents \citep{bretto2013hypergraph}.
Hypergraphs have been employed in areas such as chemistry \citep{konstantinova2001application}, point cloud processing \citep{zhang2020hypergraph}, and sensor network analysis \citep{yi2020hypergraph}.
Most existing works propose message passing mechanisms over hypergraphs with a pre-defined topology. 
In contrast, we propose an effective mechanism to infer the hypergraph topology for group-aware relational reasoning in a data-driven manner. Moreover, we propose a hypergraph evolution mechanism to enable dynamic reasoning about both pair-wise and group-wise relations.
It is also necessary to encourage smooth evolution of the relational structures and adjust their sparsity flexibly for different applications. 

The main contributions of this paper are as follows:
\begin{itemize}
	\item We propose a group-aware relational reasoning approach by inferring dynamically evolving latent interaction graphs and hypergraphs based on observations, and demonstrate its effectiveness for multi-agent trajectory prediction. We use a flexible mechanism to learn the construction of hyperedges, which are trained end-to-end. 
	\item We propose effective mechanisms to regularize the smoothness of relation evolution over time and the sparsity of learned graphs or hypergraphs, which not only enhances training stability and robustness but also reduces prediction error.
	\item We validate the proposed model on both crowd simulations and trajectory prediction benchmarks. Our EvolveHypergraph can infer explainable, evolving hypergraph structures for group-aware relational reasoning and achieve state-of-the-art prediction performance. 
\end{itemize}

\vspace{-0.1cm}
\section{Related work}
\vspace{-0.1cm}
The objective of trajectory prediction is to forecast a sequence of future states based on the historical observations, which has been widely studied especially in dense scenarios with highly interactive entities (e.g., pedestrians \citep{rudenko2020human,kothari2021human}, vehicles \citep{huang2022survey}, sports players \citep{kipf2018neural,li2020evolvegraph,graber2020dynamic}).
Traditional approaches such as physics-based models \citep{helbing1995social,mehran2009abnormal}, heuristics-based models \citep{berg2011reciprocal,nilsson2015rule}, or probabilistic graphical models \citep{wang2011trajectory,schulz2018multiple}) achieved satisfactory performance in simple scenarios where agents mostly move independently.
Recent advances in deep learning models have enhanced the capability of capturing not only the motion patterns of individuals but also the complex interactions between agents, which improve prediction accuracy by a large margin \citep{alahi2016social,gupta2018social,bisagno2018group,mohamed2020social,huang2019stgat,kosaraju2019social,li2019conditional,salzmann2020trajectron++,mangalam2020not,yu2020spatio,zhao2021tnt,giuliari2021transformer,yuan2021agentformer,zhou2022grouptron,xu2022groupnet}.

The relational learning and reasoning module is a crucial component in multi-agent trajectory prediction \citep{rudenko2020human,huang2022survey}. It also shows effectiveness in social robot navigation \citep{mavrogiannis2021core}, physical dynamics modeling \citep{sanchez2020learning,battaglia2016interaction}, human-object interaction detection \citep{gao2020drg}, group activity recognition \citep{perez2022skeleton}, and community detection in various types of networks \citep{su2022comprehensive}.
The performance of these tasks highly depends on the learned relations or inferred interactions between different entities.
Traditional statistical relational learning approaches were proposed to infer the dependency between random variables based on statistics and logic, which are difficult to leverage informative high-dimensional data (e.g., images/videos) for inferring complex relations \citep{koller2007introduction}.
Recently, various deep learning techniques have been investigated to model the relations or interactions such as pooling layers \citep{deo2018convolutional,alahi2016social}, attention mechanisms \citep{niu2021review,vemula2018social}, transformers \citep{yu2020spatio,giuliari2021transformer,yuan2021agentformer}, and graph neural networks \citep{battaglia2018relational,kipf2018neural,li2020evolvegraph,graber2020dynamic,li2021rain}, which can extract flexible relational features used in downstream tasks by aggregating the information of individual entities.
In constrast with most existing graph-based methods, which only use edges to model pair-wise relations between agents, we propose to adopt effective hyperedges to additionally capture dynamic group-wise relations by inferring evolving hypergraphs.

\vspace{-0.1cm}
\section{EvolveHypergraph} 
\vspace{-0.1cm}

We denote a set of agents' trajectories as $\mathbf{X}_{1:T} = \{\mathbf{x}^i_{1:T}, T=T_\text{h}+T_\text{f},i=1,\ldots,N\}$ where $N$ represents the number of involved agents, $T_\text{h}$ and $T_\text{f}$ represent the history and future prediction horizons, respectively.
More specifically, we have $\mathbf{x}^i_t = (x^i_t,y^i_t)$ as the state of agent $i$, where $(x,y)$ represents the 2D coordinate.
The final objective of probabilistic trajectory prediction is to estimate the conditional distribution $p(\mathbf{X}_{T_\text{h}+1:T_\text{h}+T_\text{f}}\mid \mathbf{X}_{1:T_\text{h}})$.
As an intermediate step, we also aim to infer the underlying relational structures in the form of dynamic graphs (i.e., pair-wise relations) and hypergraphs (i.e., group-wise relations), which may evolve over time and largely influence the predicted trajectories.
The long-term prediction can be obtained by iterative short-term predictions with a horizon of $T_\text{p}$ ($\leq T_\text{f}$), which depends on the evolving relations.

\vspace{-0.2cm}
\subsection{Method overview}
\vspace{-0.1cm}
\begin{wrapfigure}[12]{R}{0.5\textwidth}
	\centering
	\vspace{-0.7cm}
	\includegraphics[width=0.5\textwidth]{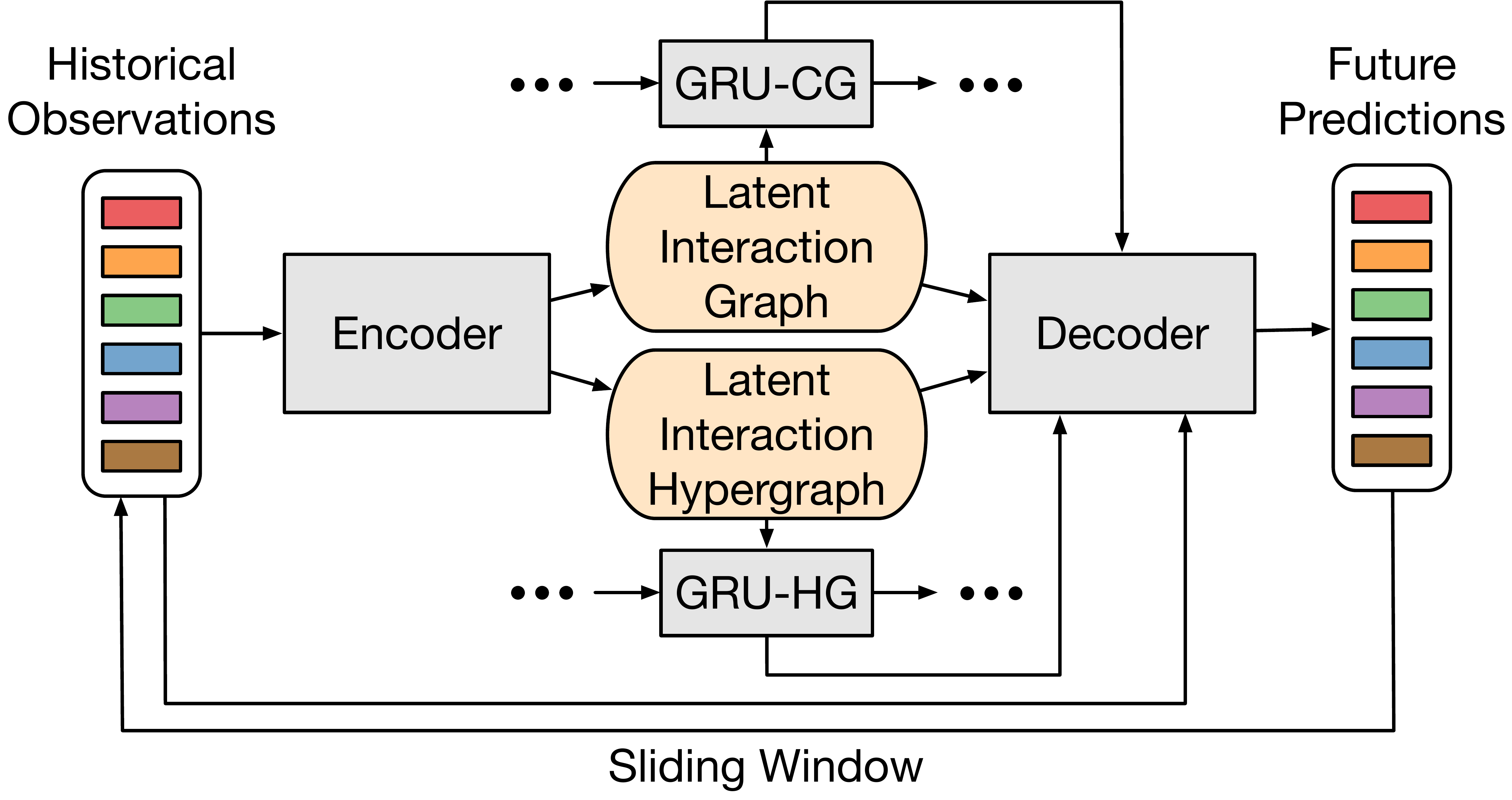}
	\vspace{-0.6cm}
	\caption{A diagram of EvolveHypergraph with dynamic relational reasoning.}
	\label{fig:model_diagram_dynamic}
\end{wrapfigure}

\begin{figure}[!tbp]
	\centering
	\includegraphics[width=0.90\textwidth]{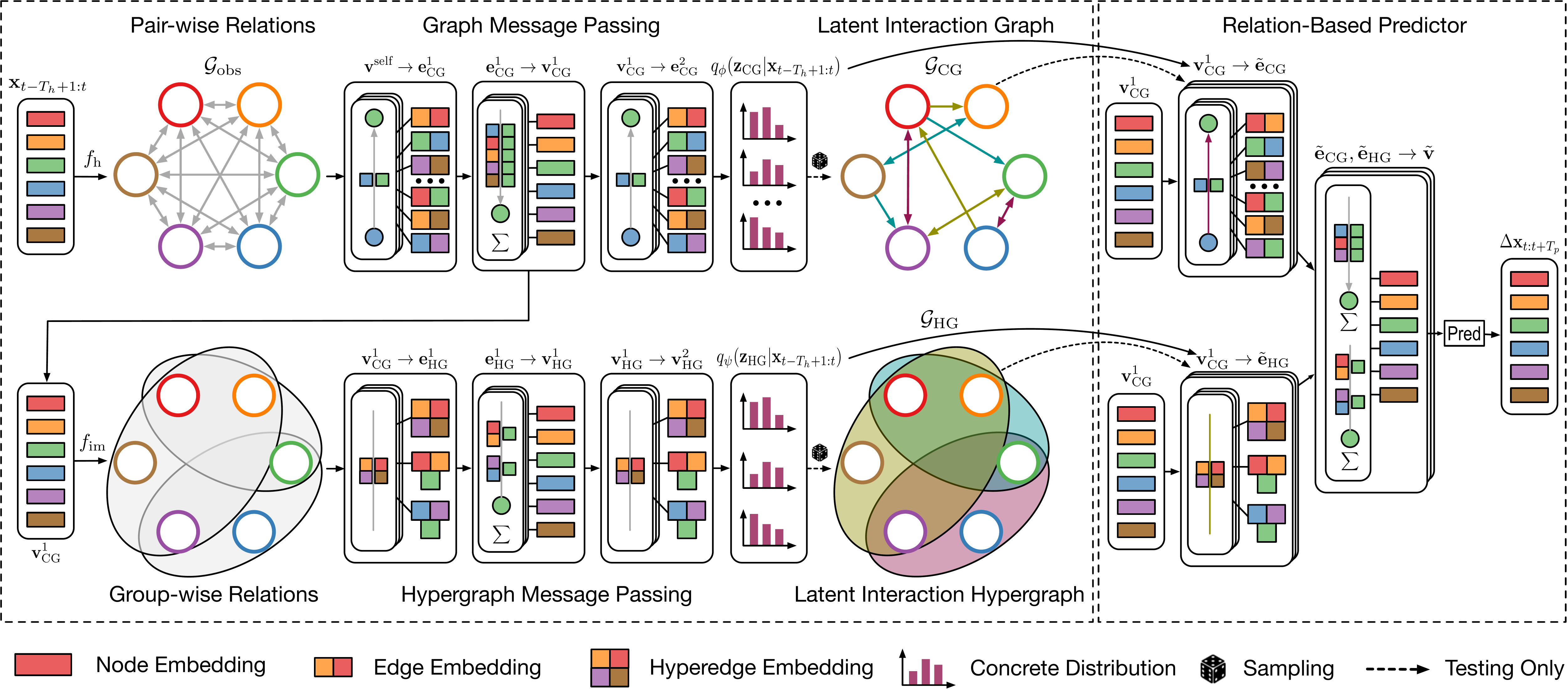}
	\vspace{-0.2cm}
	\caption{A diagram of EvolveHypergraph with static relational reasoning in a single prediction period, which consists of an encoder for relational reasoning and a decoder to generate prediction hypotheses. Different colors of edges in the latent interaction graph and hyperedges (i.e., ellipses) in the latent interaction hypergraph indicate different relation types. Best viewed in color.}
	\vspace{-0.3cm}
	\label{fig:model_diagram_static}
\end{figure}

Figure \ref{fig:model_diagram_dynamic} shows the major components of our EvolveHypergraph approach, which follows an encoder-decoder architecture with an iterative sliding window to enable dynamic relational reasoning for long-term prediction. 
The encoder infers relation graphs and hypergraphs in the latent space based on historical state observations without direct supervisions on relations. 
The decoder leverages the inferred latent relational structures to generate the distribution of future trajectory hypotheses. 
Figure \ref{fig:model_diagram_static} shows the operations in the encoder and decoder within a single prediction period.
Moreover, we present a dynamic reasoning mechanism to capture the evolution of both pair-wise and group-wise relations with uncertainty.

Similar to the formulation in \citep{li2020evolvegraph}, we employ a fully-connected observation graph $\mathcal{G}_\text{obs} = \{ \mathcal{V}_\text{obs}, \mathcal{E}_\text{obs}\}$ to represent the involved agents, where $\mathcal{V}_\text{obs} = \{ \mathbf{v}_i, i\in \{1,\ldots,N\} \}$ and $\mathcal{E}_\text{obs} = \{ \mathbf{e}_{ij}, i,j\in \{1,\ldots,N\} \}$. $\mathbf{v}_i$ and $\mathbf{e}_{ij}$ denote the attribute of node $i$ and the attribute of the directed edge from node $j$ to node $i$, respectively.
Each agent node has two types of attributes: a \textit{self-attribute} which encodes its own state information and a \textit{social-attribute} which encodes the collection of others' information.
The details of the encoder and decoder are elaborated in the following.

\vspace{-0.2cm}
\subsection{Encoder: group-aware relational reasoning}
\vspace{-0.1cm}

As is shown in Figure \ref{fig:model_diagram_static}, the encoder consists of two parallel branches: \textit{pair-wise} relational reasoning using a conventional graph ($\mathcal{G}_\text{CG}$) and \textit{group-wise} relational reasoning using a hypergraph ($\mathcal{G}_\text{HG}$), which are correlated via the node attributes obtained after a round of message passing across the observation graph $\mathcal{G}_\text{obs}$.

\textbf{Computing pair-wise edge features} The self-attribute $\mathbf{v}^{\text{self}}_i$ and social attribute $\mathbf{v}^{\text{social}}_i$ of agent $i$ are 
{
\small
\begin{align}
	\mathbf{v}^{\text{self}}_i = \ f_\text{h} \left(\mathbf{x}^i_{t-T_h+1:t}\right), \ \mathbf{e}^{1}_{\text{CG},ij} = \ f^1_{\text{CG},e}\left(\left[\mathbf{v}^{\text{self}}_i, \ \mathbf{v}^{\text{self}}_j\right]\right), \
	\mathbf{v}^{\text{social}}_i = \ f^1_{\text{CG},v} \left( \sum_{i\ne j} \mathbf{e}^{1}_{\text{CG},ij}\right),
\end{align}
}%
where $f_\text{h}(\cdot)$ denotes the history embedding function, $\mathbf{e}^{1}_{\text{CG},ij}$, $f^1_{\text{CG},e}(\cdot)$ and $f^1_{\text{CG},v}(\cdot)$ denote the edge attribute, edge and node update functions during the first round of message passing across the conventional graph, respectively.
Then we can obtain the complete node attribute of agent $i$ and conduct a second round of edge update by
{\small
\begin{align}
	\mathbf{v}^1_{\text{CG},i} = \ \left[\mathbf{v}^{\text{self}}_i, \ \mathbf{v}^{\text{social}}_i\right],  \quad
	\mathbf{e}^{2}_{\text{CG},ij} = \ f^2_{\text{CG},e}\left(\left[\mathbf{v}^1_{\text{CG},i}, \ \mathbf{v}^1_{\text{CG},j}\right]\right).
\end{align}
}%
\textbf{Computing hyperedge features} 
We infer the topology of the hypergraph $\mathcal{G}_\text{HG}$ in the form of incidence matrix $I_\text{HG}$ by $I_\text{HG} = f_\text{im} (\mathbf{V}^1_\text{CG})$.
More specifically, we first obtain a probabilistic incidence matrix by $I_\text{PIM} = f_\text{PIM} (\mathbf{V}^1_\text{CG})$
where $I_{\text{PIM},im}$ is the probability of node $i$ being connected by hyperedge $\mathcal{H}_m$,
the $i$-th row of the node attribute matrix $\mathbf{V}^1_\text{CG}$ is $\mathbf{v}^1_{\text{CG},i}$, and $f_{\text{PIM}}(\cdot)$ is a multi-layer perceptron (MLP).
Then, we obtain the incidence matrix $I_{\text{HG}}$ by sampling the value of $I_{\text{HG},im}$ (1 or 0) from a Gumbel-Softmax distribution \citep{maddison2017concrete} defined by $I_{\text{PIM},im}$.
We do not sample from a Bernoulli distribution because of the backpropagation issue \citep{maddison2017concrete}.
In summary, we have $f_{\text{im}}(\mathbf{V}^1_\text{CG}):= \text{Gumbel-Softmax}(f_\text{PIM} (\mathbf{V}^1_\text{CG}))$.
Note that we need to set the maximum number of hyperedges denoted as $M$ according to specific applications.
Based on the inferred $I_\text{HG}$, the message passing across the hypergraph is given by
{\small
\begin{align}
	\mathbf{e}^{1}_{\text{HG},m} = \ f^1_{\text{HG},e}\left(\sum_{i\in \mathcal{H}_m}\mathbf{v}^1_{\text{CG},i} \right), 
	\mathbf{v}^1_{\text{HG},i} = \ f^1_{\text{HG},v} \left( \sum_{m:i\in \mathcal{H}_m} \mathbf{e}^{1}_{\text{HG},m} \right), 
	\mathbf{e}^{2}_{\text{HG},m} = \ f^2_{\text{HG},e}\left(\sum_{i\in \mathcal{H}_m}\mathbf{v}^1_{\text{HG},i}\right),
\end{align}
}%
where $\mathbf{e}^{1}_{\text{HG},m}$ denotes the attribute of hyperedge $\mathcal{H}_m$ ($m \in \{1,\ldots,M\}$), $f^1_{\text{HG},e}(\cdot)/f^2_{\text{HG},e}(\cdot)$ and $f^1_{\text{HG},v}(\cdot)$ denote the hyperedge and node update functions during the message passing across the hypergraph, respectively.
Our message passing mechanism is adapted from \citep{huang2021unignn} and \citep{chien2021you}.

\textbf{Inferring relation patterns} The interaction conventional graph $\mathcal{G}_\text{CG}$ and hypergraph $\mathcal{G}_\text{HG}$ represent relation patterns with a distribution of edge / hyperedge types, which are inferred based on the edge / hyperedge attributes.
We denote the number of possible edge / hyperedge types as $L_\text{CG}$ and $L_\text{HG}$ which can be adjusted for different applications.
Since not all pairs of agents have direct relations, we set the first type as ``no edge'' which prevents message passing through the corresponding edges and can be used to regularize the sparsity of relations.
Formally, we formulate a classification problem which is to infer the edge / hyperedge (i.e., relation) types based on $\mathbf{e}^{2}_{\text{CG},ij}$ and $\mathbf{e}^{2}_{\text{HG},m}$ by
{\small
\begin{align}
	\mathbf{z}_{\text{CG},ij} =& \ \text{Softmax}\left(\left(\mathbf{e}^2_{\text{CG},ij}+\mathbf{g}\right)/\tau\right), \ i,j\in \{1,\ldots,N\},\\
	\mathbf{z}_{\text{HG},m} =& \ \text{Softmax}\left(\left(\mathbf{e}^2_{\text{HG},m}+\mathbf{g}\right)/\tau\right), \ m\in \{1,\ldots,M\},
\end{align}
}%
where $\mathbf{g}$ denotes a vector of i.i.d. samples from the Gumbel$(0,1)$ distribution and $\tau$ denotes the temperature coefficient which controls the smoothness of samples \citep{maddison2017concrete}.
We can use the reparametrization trick to get gradients from this approximation, which is adopted from NRI \citep{kipf2018neural} and EvolveGraph \citep{li2020evolvegraph}.
The latent variables $\mathbf{z}_{\text{CG},ij}$ and $\mathbf{z}_{\text{HG},m}$ are multinomial random vectors, which indicate the relation types for a certain edge / hyperedge.
More formally, we have 
\begin{align}
    \mathbf{z}_{\text{CG},ij} = \left[z_{\text{CG},ij,1},\ldots,z_{\text{CG},ij,L_\text{CG}}\right], \quad
    \mathbf{z}_{\text{HG},m} = \left[z_{\text{HG},m,1},\ldots,z_{\text{HG},m, L_\text{HG}}\right],
\end{align}
where $L_\text{CG}$ and $L_\text{CG}$ denote the number of edge types and hyperedge types, respectively.
$z_{\text{CG},ij,r} \ (1\leq r \leq L_\text{CG})$ and $z_{\text{HG},m,s} \ (1\leq s \leq L_\text{HG})$ denote the probability of a certain edge / hyperedge belongs to a certain type, respectively.
In short, the operations in the encoder can be summarized as
{\small
\begin{align}
	q_\text{CG}\left(\mathbf{z}_{\text{CG},\beta}\mid \mathbf{X}_{t-T_h+1:t}\right), \ q_\text{HG}\left(\mathbf{z}_{\text{HG},\beta}\mid \mathbf{X}_{t-T_h+1:t}\right) = \ \text{Encoder}\left(\mathbf{X}_{t-T_h+1:t}\right).
\end{align}
}%

\subsection{Decoder: relation-based predictor}
\textbf{Relational feature aggregation} To enable dynamic relational reasoning, the whole prediction process consists of multiple recurrent prediction period with time horizon $T_\text{p}$ during which the relation graphs / hypergraphs are re-inferred.
In a single prediction period, the decoder is designed to predict the distribution of future trajectories conditioned on the historical observations or the prediction hypotheses in the last prediction period as well as the relation graphs / hypergraphs inferred by the encoder, which is written as $p\left(\mathbf{X}_{t+1:t+T_\text{p}}\mid \mathbf{X}_{t-T_\text{h}+1:t}, \mathcal{G}_\text{CG}, \mathcal{G}_\text{HG}\right)$.
The detailed operations are:
{\small
\begin{align}
	\Tilde{\mathbf{e}}_{\text{CG},ij} = \sum^{L_\text{CG}}_{l=1} z_{\text{CG},ij,l} \Tilde{f}^l_{\text{CG},e} &\left(\left[\mathbf{v}^1_{\text{CG},i},\ \mathbf{v}^1_{\text{CG},j}\right]\right), \
	\Tilde{\mathbf{e}}_{\text{HG},m} = \sum^{L_\text{HG}}_{l'=1} z_{\text{HG},m,l'} \Tilde{f}^{l'}_{\text{HG},e} \left( \sum_{i\in \mathcal{H}_m}\mathbf{v}^1_{\text{CG},i} \right), \label{eq:agg}\\
	\Tilde{\mathbf{v}}_{i} =& \ \Tilde{f}_v \left(\left[ \sum_{i\ne j}\Tilde{\mathbf{e}}_{\text{CG},ij}, 
	\sum_{i\in \mathcal{H}_m} \Tilde{\mathbf{e}}_{\text{HG},m}\right]\right), \label{eq:10}
\end{align}
}%
where the node attributes $\mathbf{v}^1_{\text{CG},i}$ obtained in the encoding process are leveraged to obtain the aggregated edge / hyperedge attributes $\Tilde{\mathbf{e}}_{\text{CG},ij}$ and $\Tilde{\mathbf{e}}_{\text{HG},m}$ in equation (\ref{eq:agg}).
$\Tilde{f}^l_{\text{CG},e}(\cdot)$ and $\Tilde{f}^{l'}_{\text{HG},e}(\cdot)$ denote the edge update functions corresponding to a certain edge / hyperedge type in $\mathcal{G}_\text{CG}$ and $\mathcal{G}_\text{HG}$, respectively. 
$\Tilde{f}_v$ denotes the node update function in the decoding process.

\textbf{Predicting trajectory distributions} To capture the uncertainty of future trajectories, the predicted distribution of displacement $\Delta\mathbf{x}^i_t$ at each time step is designed to be a Gaussian distribution with an equal constant covariance. 
The output function $\Tilde{f}_\text{output}(\cdot)$ takes in the updated node attribute $\Tilde{\mathbf{v}}_i$ and output a sequence of the mean of the Gaussian kernel of displacement $\Delta\mathbf{x}^i_{t:t+T_\text{p}-1}$ for agent $i$.
Then we can iteratively calculate the mean of the coordinates of agent $i$ in the future by
{\small
\begin{align}
	\hat{\bm{\mu}}^{i}_{t+1} = \hat{\mathbf{x}}^i_t + \Delta\hat{\mathbf{x}}^i_t \ = \ \hat{\mathbf{x}}^i_t + \Tilde{f}_{\text{output},t}\left( \Tilde{\mathbf{v}}_i\right).
\end{align}
}%
Then the predicted conditional distribution is obtained by
{\small
\begin{align}
	p\left(\hat{\mathbf{x}}_{t+1}^{i}\mid \mathbf{z}_\text{CG}, \mathbf{z}_\text{HG}, \mathbf{x}_{t-T_h+1:t}^{i}\right) = \  \mathcal{N}\left(\hat{\mathbf{x}}_{t+1}^{i}\mid \hat{\bm{\mu}}^{i}_{t+1}, \sigma^2 \mathbf{I}\right),
\end{align}
}%
where $\sigma^2$ denotes the constant variance of Gaussian kernels, and $\mathbf{I}$ denotes the identity matrix.

\subsection{Evolution of relations}

We generalize the graph evolving mechanism proposed in \citep{li2020evolvegraph} to the dynamic inference of hypergraphs to capture evolving relations and the uncertainty of future trajectories.
The detailed operations in the overall diagram in Figure \ref{fig:model_diagram_dynamic} are written as follows:
{\small
\begin{align}
	q_\text{CG}\left(\mathbf{z}_{\text{CG},\beta}\mid \mathbf{X}_{1+\beta\tau:T_h+\beta\tau}\right),& \ q_\text{HG}\left(\mathbf{z}_{\text{HG},\beta}\mid \mathbf{X}_{1+\beta\tau:T_h+\beta\tau}\right) = \ \text{Encoder}\left(\mathbf{X}_{1+\beta\tau:T_h+\beta\tau}\right), \\
	q_\text{CG}\left(\mathbf{z}'_{\text{CG},\beta}\mid \mathbf{X}_{1+\beta\tau:T_h+\beta\tau}\right) &= \ \text{GRU-CG}\left(q\left(\mathbf{z}_{\text{CG},\beta}\mid \mathbf{X}_{1+\beta\tau:T_h+\beta\tau}\right), \mathbf{H}_{\text{CG},\beta} \right),\\
	q_\text{HG}\left(\mathbf{z}'_{\text{HG},\beta}\mid \mathbf{X}_{1+\beta\tau:T_h+\beta\tau}\right) &= \ \text{GRU-HG}\left(q\left(\mathbf{z}_{\text{HG},\beta}\mid \mathbf{X}_{1+\beta\tau:T_h+\beta\tau}\right), \mathbf{H}_{\text{HG},\beta} \right),
\end{align}
}%
where $\tau$ denotes the time gap between two consecutive relational inference steps, $\beta$ denotes the index of relation graphs/ hypergraphs starting from 0, $\mathbf{H}_{\text{CG},\beta}$ and $\mathbf{H}_{\text{HG},\beta}$ denote the hidden states of recurrent units GRU-CG and GRU-HG that propagate the corresponding relational structures every $\tau$ steps, $\mathbf{z}'_{\text{CG},\beta}$ and $\mathbf{z}'_{\text{HG},\beta}$ denote the updated pair-wise and group-wise relational structures, respectively.
Finally, the future trajectory distribution is obtained by
{\small
\begin{align}
	p\left(\mathbf{X}_{T_h+\beta\tau+1:T_h+(\beta+1)\tau}\mid \mathcal{G}'_{\text{CG},\beta},\mathcal{G}'_{\text{HG},\beta},\mathbf{X}_{1:T_h},\hat{\mathbf{X}}_{T_h+1:T_h+\beta\tau}\right)& \nonumber \\  = \ \text{Decoder}\left(\mathcal{G}'_{\text{CG},\beta},\mathcal{G}'_{\text{HG},\beta},\mathbf{X}_{1:T_h},\hat{\mathbf{X}}_{T_h+1:T_h+\beta\tau}\right)&.
\end{align}
}%

\subsection{Regularization on relations: smoothness and sparsity}

We propose effective regularization techniques on the learned underlying relational structures from two aspects.
First, both pair-wise and group-wise relations tend to evolve gradually and smoothly over time.
Therefore, we propose a smoothness regularization loss that indicates the difference between two consecutive graphs or hypergraphs, which is calculated by
{\small
\begin{align}
	L_\text{SM} =& \ \alpha_\text{SM, CG}\text{KL}\left( q_\text{CG}\left(\mathbf{z}'_{\text{CG},\beta}\mid\mathbf{X}_{1+\beta\tau:T_h+\beta\tau}\right) \, || \, \ q_\text{CG}\left(\mathbf{z}'_{\text{CG},\beta+1}\mid\mathbf{X}_{1+(\beta+1)\tau:T_h+(\beta+1)\tau}\right) \right) \nonumber \\
	+& \ \alpha_\text{SM, HG}\text{KL}\left(q_\text{HG}\left(\mathbf{z}'_{\text{HG},\beta}\mid\mathbf{X}_{1+\beta\tau:T_h+\beta\tau}\right) \,||\,  \ q_\text{HG}\left(\mathbf{z}'_{\text{HG},\beta+1}\mid\mathbf{X}_{1+(\beta+1)\tau:T_h+(\beta+1)\tau}\right)\right),
\end{align}
}%
where $\alpha_\text{SM, CG}$ and $\alpha_\text{SM, HG}$ denote the corresponding coefficients and KL$(\cdot)$ denotes the Kullback–Leibler divergence.
The intuition is that the consecutive graphs / hypergraphs should have similar topologies in the form of edge / hyperedge type distributions.

Second, human interactions tends to be sparse even in a crowded scenario. They only interact with a subset of surrounding agents, which implies that not all the pairs of agents have a direct relation. 
Therefore, the learned relational structures should have an appropriate sparsity.
We propose a sparsity regularization loss based on the entropy of edge / hyperedge type distributions, which is calculated by
{\small
\begin{align}
	L_\text{SP} = \ -& \ \alpha_\text{SP, CG}q_\text{CG}\left(\mathbf{z}'_{\text{CG},\beta}\mid\mathbf{X}_{1+\beta\tau:T_h+\beta\tau}\right) \log q_\text{CG}\left(\mathbf{z}'_{\text{CG},\beta}\mid\mathbf{X}_{1+\beta\tau:T_h+\beta\tau}\right) \nonumber \\
	-& \ \alpha_\text{SP, HG} q_\text{HG}\left(\mathbf{z}'_{\text{HG},\beta}\mid\mathbf{X}_{1+\beta\tau:T_h+\beta\tau}\right) \log q_\text{HG}\left(\mathbf{z}'_{\text{HG},\beta}\mid\mathbf{X}_{1+\beta\tau:T_h+\beta\tau}\right),
\end{align}
}%
where $\alpha_\text{SP, CG}$ and $\alpha_\text{SP, HG}$ denote the corresponding coefficients.

\subsection{Training strategy and loss function}
Since the inference of hyperedges highly depends on the intermediate node attributes $\mathbf{V}^1_\text{CG}$ obtained after the message passing across the observation graph $\mathcal{G}_\text{obs}$, we adopt a warm-up training stage during which only the modules related to the pair-wise relational reasoning is trained. This training strategy provides a good initialization for the inference of hyperedges, which proves to accelerate convergence and improve the final performance. 
The complete loss at the formal training stage consists of a reconstruction loss, KL divergence losses, and regularization losses, which are calculated by
{\small
\begin{align}
	L = L_\text{Rec} &\ + \ L_\text{KL} + L_\text{SM} + L_\text{SP}, \quad
	L_\text{Rec} = \sum_{i=1}^{N}\sum_{t=T_\text{h}+1}^{T_\text{h}+T_\text{f}} \|\mathbf{x}^i_t - \bm{\mu}^i_t\|^2, \\
	L_\text{KL} =& \  \alpha_\text{KL, CG}\text{KL}\left( q_\text{CG}\left(\mathbf{z}'_{\text{CG},\beta}\mid\mathbf{X}_{1+\beta\tau:T_h+\beta\tau}\right) \,||\, \ p_\text{CG}\left(\mathbf{z}'_{\text{CG},\beta}\right)\right)  \nonumber \\ 
	+& \ \alpha_\text{KL, CG}\text{KL}\left( q_\text{HG}\left(\mathbf{z}'_{\text{HG},\beta}\ | \ \mathbf{X}_{1+\beta\tau:T_h+\beta\tau}\right)\,||\, \ p_\text{HG}\left(\mathbf{z}'_{\text{HG},\beta}\right)\right),
\end{align}
}%
where $p_\text{CG}\left(\mathbf{z}'_{\text{CG},\beta}\right)$ and $p_\text{HG}\left(\mathbf{z}'_{\text{HG},\beta}\right)$ denotes the prior uniform categorical distributions, and $\alpha_\text{KL, CG}$ and $\alpha_\text{KL, HG}$ denote the corresponding coefficients. The whole model can be trained end-to-end.

\section{Experiments}
We tested the proposed approach EvolveHypergraph on a crowd simulation dataset generated by a group-based simulator modified from \citep{SFMG} and three standard benchmark datasets for human trajectory prediction: ETH/UCY \citep{pellegrini2009you,lerner2007crowds} and SDD \citep{robicquet2016learning} datasets (real-world pedestrians), and NBA SportVU dataset\footnote{\url{https://github.com/linouk23/NBA-Player-Movements}} (team-based sports).
The simulation / dataset details, implementation details, and additional experimental results can be found in the supplementary materials.

\subsection{Evaluation metrics and baselines}
To evaluate the trajectory prediction performance on both synthetic and benchmark datasets, we adopted two standard evaluation metrics same as EvolveGraph \citep{li2020evolvegraph}: minADE$_{20}$ and minFDE$_{20}$, which indicates the minimum error among $20$ trajectory samples over the whole prediction horizon and at the final time step, respectively.
For benchmark datasets, we compared our approach with several state-of-the-art baselines that adopt a certain technique to model interactions such as latent graph inference (i.e., NRI \citep{kipf2018neural}, DNRI \citep{graber2020dynamic}, EvolveGraph \citep{li2020evolvegraph}), attention mechanisms (i.e., STGAT \citep{huang2019stgat}), social pooling (i.e., PECNet \citep{mangalam2020not}) and graph message passing (i.e., Trajectron++ \citep{salzmann2020trajectron++}).
Since we can obtain the ground truth group-wise relations in the crowd simulation, we also evaluated the relational reasoning performance by comparing the inferred and true incidence matrices.
Moreover, we conducted an ablation study to demonstrate the effectiveness of each model component.

\begin{figure}[!tbp]
	\centering
	\includegraphics[width=\textwidth,height=7.8cm]{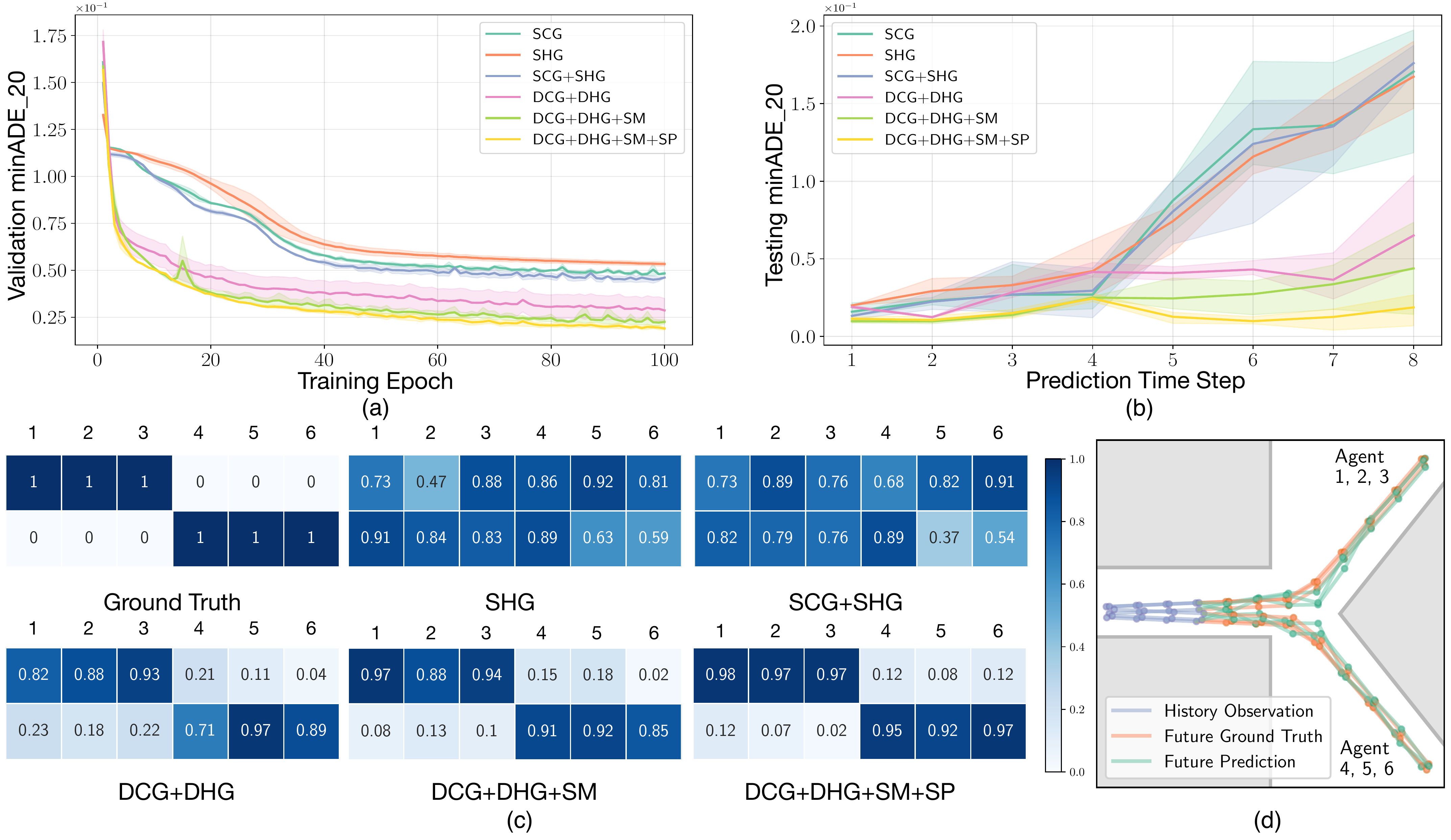}
	\vspace{-0.5cm}
	\caption{The visualizations of results on the crowd simulation dataset. (a) minADE$_{20}$ vs. training epochs on the validation dataset. The solid lines indicates the average error and the shaded areas indicate variance. (b) minADE$_{20}$ vs. prediction horizon on the testing dataset. (c) The incidence matrix of the hypergraphs inferred by different model settings. Each row represents a hyperedge and each column represents a certain agent. The numbers indicate the probability of the agent belonging to the corresponding hyperedge (i.e., group). The hyperedges of SHG and SCG+SHG shown in the figure are inferred at the first prediction step while the others are inferred at a step after the group evolution. (d) The predicted trajectories based on a sampled combination of graph and hypergraph inferred by the DCG+DHG+SM+SP model. The agent index corresponds to the column index in the incidence matrices in (c). Best viewed in color.} 
	\vspace{-0.3cm}
	\label{fig:crowd_simulation}
\end{figure}

\subsection{Quantitative analysis}

\noindent
\textbf{ETH / UCY and SDD datasets}: 
These two datasets were collected in real-world traffic scenarios where pedestrians move on the street or in the university campus. Some scenarios are crowded and full of pair-wise and group-wise interactions between pedestrians who naturally form groups and move together. 
We adopted the standard setting for training and testing our model to predict trajectories in the future 4.8s (12 time steps) based on 3.2s (8 time steps) historical observations.
Our approach reduces minADE$_{20}$ / minFDE$_{20}$ by 5.3\% / 9.8\% and 4.0\% / 5.7\% on the ETH /UCY and SDD datasets compared with the previous strongest baselines, respectively.

\noindent
\textbf{NBA dataset}:
This dataset was collected during real NBA games where the basketball players have highly intensive interactions and agile motion patterns.
Different from the pedestrians in a group that behave similarly, a group of basketball players tend to include adversarial agents from different teams, which creates even more complex group behaviors.
Owing to the effective group-wise relational reasoning module, our approach reduces minADE$_{20}$ / minFDE$_{20}$ by 26.2\% / 34.1\% on the NBA dataset compared with the previous strongest baseline.

\vspace{-0.1cm}
\subsection{Ablative analysis}
\vspace{-0.1cm}
We conducted an ablation study on both the synthetic simulation and benchmark datasets to demonstrate the effectiveness of model components in our approach from multiple aspects.
we presented quantitative results of five ablation settings in addition to our complete model in Figure \ref{fig:crowd_simulation}(a)(b), Table \ref{tab:ETH-UCY}, Table \ref{tab:SDD}, and Table \ref{tab:NBA}. The ablation settings are introduced below.
\begin{itemize}
	\item SCG: The model with only \textit{pair-wise} static relational reasoning;
	\item SHG: The model with only \textit{group-wise} static relational reasoning;
	\item SCG+SHG: The model with both \textit{pair-wise} and \textit{group-wise} static relational reasoning;
	\item DCG+DHG: The model with both pair-wise and group-wise \textit{dynamic} relational reasoning;
	\item DCG+DHG+SM: The DCG+DHG model with additional \textit{smoothness} regularization loss on evolving graphs / hypergraphs;
	\item DCG+DHG+SM+SP: The DCG+DHG+SM model with additional \textit{sparsity} regularization loss on evolving graphs / hypergraphs.
\end{itemize}

\begin{table*}[!tbp]
	\setlength{\tabcolsep}{1mm}
	\caption{minADE$_{20}$ / minFDE$_{20}$ (meters) on the ETH-UCY dataset.}
	\vspace{-0.cm}
	\fontsize{8}{8}\selectfont
	\resizebox{\textwidth}{!}{
		\begin{tabular}{m{0.9cm}<{\centering}|m{1.3cm}<{\centering} m{1.3cm}<{\centering} m{1.3cm}<{\centering} m{1.3cm}<{\centering} m{1.5cm}<{\centering} m{1.5cm}<{\centering} | m{1.3cm}<{\centering} m{1.3cm}<{\centering} m{1.3cm}<{\centering} m{1.3cm}<{\centering} m{1.5cm}<{\centering} m{1.5cm}<{\centering}}
			\toprule
			\midrule
			\multirow{3}*{\shortstack[lb]{}} 
			& \multicolumn{6}{c|}{Baseline Methods} & \multicolumn{6}{c}{\textbf{EvolveHypergraph (Ours)}} \\
			\cline{2-13} 
			\vspace{0.8cm}
			Scene & NRI \citep{kipf2018neural}  &  DNRI \citep{graber2020dynamic} & EvolveGraph \citep{li2020evolvegraph} & STGAT \citep{huang2019stgat} &PECNet \citep{mangalam2020not}  & Trajectron++ \citep{salzmann2020trajectron++}  & SCG & SHG & SCG + SHG & DCG + DHG & DCG + DHG + SM & \textbf{DCG + DHG + SM + SP} \\
			\midrule 
			ETH   & 0.48 / 0.89 & 0.32 / 0.59 & \textbf{0.31} / \textbf{0.58} & 0.65 / 1.12 & 0.54 / 0.87 & 0.39 / 0.83 & 0.40 / 0.72 & 0.44 / 0.78 & 0.37 / 0.68 & 0.35 / 0.66 & 0.34 /0.64 & 0.32 / 0.61 \\
			HOTEL & 0.20 / 0.38 & 0.17 / 0.33 & 0.18 / 0.35 & 0.35 / 0.66 & 0.18 / 0.24 & 0.12 / 0.21 & 0.16 / 0.28 & 0.17 / 0.29 & 0.14 / 0.25 & 0.14 / 0.24 & 0.12 / 0.22 & \textbf{0.11} / \textbf{0.20} \\
			UNIV  & 0.31 / 0.67 & 0.26 / 0.55 & 0.24 / 0.51 & 0.52 / 1.10 & 0.35 / 0.60 & 0.20 / 0.44 & 0.22 / 0.53 & 0.25 / 0.57 & 0.21 / 0.50 & 0.20 / 0.48 & 0.19 / 0.45 & \textbf{0.18} / \textbf{0.43} \\
			ZARA1 & 0.19 / 0.39 & 0.17 / 0.35 & 0.16 / 0.31 & 0.34 / 0.69 & 0.22 / 0.39 & \textbf{0.15} / 0.33 & 0.21 / 0.38 & 0.24 / 0.41 & 0.20 / 0.35 & 0.19 / 0.33 & 0.17 / 0.32 & 0.16 / \textbf{0.30} \\
			ZARA2 & 0.21 / 0.37 & 0.15 / 0.30 & 0.17 / 0.32 & 0.29 / 0.60 & 0.17 / 0.30 & \textbf{0.11} / 0.25 & 0.19 / 0.30 & 0.20 / 0.31 & 0.17 / 0.27 & 0.16 / 0.26 & 0.15 / 0.24 & 0.13 / \textbf{0.22} \\
			\midrule
			AVG   & 0.24 / 0.49 & 0.22 / 0.42 & 0.21 / 0.43 & 0.43 / 0.83 & 0.29 / 0.48 & 0.19 / 0.41 & 0.23 / 0.51 & 0.26 / 0.55 & 0.22 / 0.48 & 0.20 / 0.44 & 0.19 / 0.43 & \textbf{0.18} / \textbf{0.37} \\
			\bottomrule
		\end{tabular}
	}
	\vspace{-0.5cm}
	\label{tab:ETH-UCY}
\end{table*}

\begin{table*}[!tbp]
	\setlength{\tabcolsep}{1mm}
	\caption{minADE$_{20}$ / minFDE$_{20}$ (pixels) on the SDD dataset.}
	\vspace{-0.cm}
	\fontsize{8}{8}\selectfont
	\resizebox{\textwidth}{!}{
		\begin{tabular}{m{0.9cm}<{\centering}|m{1.3cm}<{\centering} m{1.3cm}<{\centering} m{1.3cm}<{\centering} m{1.3cm}<{\centering} m{1.3cm}<{\centering} m{1.5cm}<{\centering} | m{1.3cm}<{\centering} m{1.3cm}<{\centering} m{1.3cm}<{\centering} m{1.3cm}<{\centering} m{1.5cm}<{\centering} m{1.5cm}<{\centering}}
			\toprule
			\midrule
			\multirow{3}*{\shortstack[lb]{}} 
			& \multicolumn{6}{c|}{Baseline Methods} & \multicolumn{6}{c}{\textbf{EvolveHypergraph (Ours)}} \\
			\cline{2-13} 
			\vspace{0.8cm}
			Time & NRI \citep{kipf2018neural}  &  DNRI \citep{graber2020dynamic} & EvolveGraph \citep{li2020evolvegraph} & STGAT \citep{huang2019stgat} & PECNet \citep{mangalam2020not}  & Trajectron++ \citep{salzmann2020trajectron++} & SCG & SHG & SCG + SHG & DCG + DHG & DCG + DHG + SM & \textbf{DCG + DHG + SM + SP} \\
			\midrule 
			4.8s & 19.7 / 33.1 & 15.2 / 23.8 & 13.9 / 22.9 & 18.8 / 31.3 & 9.9 / 15.8 & 19.3 / 32.7 & 16.9 / 28.8 & 19.3 / 32.4 & 14.3 / 23.9 & 11.4 / 19.2 & 9.9 / 16.1 & \textbf{9.5} / \textbf{14.9} \\
			\bottomrule
		\end{tabular}
	}
	\vspace{-0.5cm}
	\label{tab:SDD}
\end{table*}

\begin{table*}[!tbp]
	\setlength{\tabcolsep}{1mm}
	\caption{minADE$_{20}$ / minFDE$_{20}$ (meters) on the NBA dataset.}
	\vspace{-0.cm}
	\fontsize{8}{8}\selectfont
	\resizebox{\textwidth}{!}{
		\begin{tabular}{m{0.9cm}<{\centering}|m{1.3cm}<{\centering} m{1.3cm}<{\centering} m{1.3cm}<{\centering} m{1.3cm}<{\centering} m{1.3cm}<{\centering} m{1.5cm}<{\centering} | m{1.3cm}<{\centering} m{1.3cm}<{\centering} m{1.3cm}<{\centering} m{1.3cm}<{\centering} m{1.5cm}<{\centering} m{1.5cm}<{\centering}}
			\toprule
			\midrule
			\multirow{3}*{\shortstack[lb]{}} 
			& \multicolumn{6}{c|}{Baseline Methods} & \multicolumn{6}{c}{\textbf{EvolveHypergraph (Ours)}} \\
			\cline{2-13} 
			\vspace{0.8cm}
			Time & NRI \citep{kipf2018neural}  &  DNRI \citep{graber2020dynamic} & EvolveGraph \citep{li2020evolvegraph} & STGAT \citep{huang2019stgat} & PECNet \citep{mangalam2020not}  & Trajectron++ \citep{salzmann2020trajectron++} & SCG & SHG & SCG + SHG & DCG + DHG & DCG + DHG + SM & \textbf{DCG + DHG + SM + SP} \\
			\midrule 
			1.0s & 0.51 / 0.74 & 0.59 / 0.70 & 0.35 / \textbf{0.48} & 0.45 / 0.66 & 0.48 / 0.72 & 0.44 / 0.67 & 0.47 / 0.72 & 0.49 / 0.74 & 0.42 / 0.68 & 0.41 / 0.65 & 0.39 / 0.57 & \textbf{0.33} / 0.49\\
			2.0s & 0.96 / 1.65 & 0.93 / 1.52 & 0.66 / 0.97 & 0.87 / 1.41 & 0.91 / 1.60 & 0.79 / 1.18 & 0.92 / 1.61 & 0.95 / 1.68 & 0.86 / 1.37 & 0.77 / 1.12 & 0.68 / 0.99 & \textbf{0.63} / \textbf{0.95}\\
			3.0s & 1.42 / 2.50 & 1.38 / 2.21 & 1.15 / 1.86 & 1.28 / 2.08 & 1.33 / 2.39 & 1.51 / 2.49 & 1.39 / 2.13 & 1.44 / 2.27 & 1.24 / 2.02 & 1.07 / 1.57 & 1.02 / 1.48 & \textbf{0.93} / \textbf{1.36}\\
			4.0s & 1.86 / 3.26 & 1.78 / 2.81 & 1.64 / 2.64 & 1.69 / 2.66 & 1.73 / 3.23 & 2.09 / 3.52 & 1.80 / 2.95 & 1.91 / 3.08 & 1.69 / 2.56 & 1.40 / 2.05 & 1.32 / 1.91 & \textbf{1.21} / \textbf{1.74}\\
			\bottomrule
		\end{tabular}
	}
	\vspace{-0.3cm}
	\label{tab:NBA}
\end{table*}

\noindent
\textbf{Group-wise relational reasoning}:
We show the effectiveness of group-wise relational reasoning by comparing the SCG, SHG, and SCG+SHG models.
In all the experiments, SHG performs worse than SCG, which implies that group-wise relational reasoning through hypergraphs alone cannot sufficiently model the interactions between agents. This is reasonable because pair-wise interactions still dominate the multi-agent dynamics between the agents in the same or different groups, which cannot be entirely ignored. 
SCG+SHG achieves the best performance, which implies the effectiveness of integrating pair-wise and group-wise relational reasoning.
More specifically, SCG+SHG reduces minADE$_{20}$ by 4.4\%, 15.4\% and 6.1\% on the ETH / UCY, SDD and NBA datasets compared with SCG.
The improvement on SDD and NBA dataset are more significant than ETH / UCY since there are more group behaviors, illustrating the benefits of group-aware relational reasoning.

\noindent
\textbf{Dynamic relational reasoning}:
We show the effectiveness of dynamic relational reasoning by comparing the SCG+SHG and DCG+DHG models.
DCG+DHG outperforms the other both quantitatively and qualitatively in all the experiments. 
Particularly in the crowd simulation where the group evolves greatly, the dynamic reasoning mechanism enhances model performance in both training and testing processes.
Figure \ref{fig:crowd_simulation}(c) shows that SCG+SHG cannot predict the evolution of hyperedges which DCG+DHG can well capture in the crowd simulation.
For the benchmark datasets, DCG+DHG reduces minADE$_{20}$ by 9.1\%, 20.3\% and 17.2\% on the ETH / UCY, SDD and NBA datasets compared with SCG+SHG, illustrating its effectiveness in real-world scenarios.

\noindent
\textbf{Smoothness regularization}:
We show the effectiveness of smoothness regularization on the inferred dynamic graphs / hypergraphs by comparing the DCG+DHG and DCG+DHG+SM models, which is only applied to dynamic relational reasoning.
In Figure \ref{fig:crowd_simulation}(a) and Figure \ref{fig:crowd_simulation}(b), it is shown that DCG+DHG+SM achieves consistently better and more stable performance.
The reason is that the smoothness regularization encourages smoother evolution of graphs / hypergraphs over time and restrains abrupt changes, which leads to more stable training process and faster convergence. It also encourages the model to learn the temporal dependency between consecutively evolving relations.
Applying the smoothness regularization loss during training reduces minADE$_{20}$ by 5.0\%, 13.2\% and 5.7\% on the ETH / UCY, SDD and NBA datasets compared with DCG+DHG.

\noindent
\textbf{Sparsity regularization}:
We show the effectiveness of sparsity regularization on the inferred dynamic graphs / hypergraphs by comparing the DCG+DHG+SM and DCG+DHG+SM+SP models.
In all the experiments, DCG+DHG+SM+SP achieves the smallest prediction error among all the model settings.
Figure \ref{fig:crowd_simulation}(a) shows that the sparsity regularization further enhances the training stability indicated by the smoother curve.
Figure \ref{fig:crowd_simulation}(b) shows that DCG+DHG+SM+SP has the smallest variance in minADE$_{20}$ indicating that sparsity regularization also improves the model robustness to random initializations.
Moreover, comparing the inferred incidence matrices of the two models in Figure \ref{fig:crowd_simulation}(c), we can see that DCG+DHG+SM+SP tends to infer sparser hypergraphs which learns more distinguishable group memberships for each agent.
The potential reason for these observations is that the model is encouraged to learn to recognize sparse pair-wise relations and clear group assignments, which creates a strong inductive bias in the message passing process through the sparse relational structures.
Applying the sparsity regularization loss during training reduces minADE$_{20}$ by 5.3\%, 4.0\% and 8.3\% on the ETH / UCY, SDD and NBA datasets compared with DCG+DHG+SM.

\begin{figure}[!tbp]
	\centering
	\includegraphics[width=\textwidth]{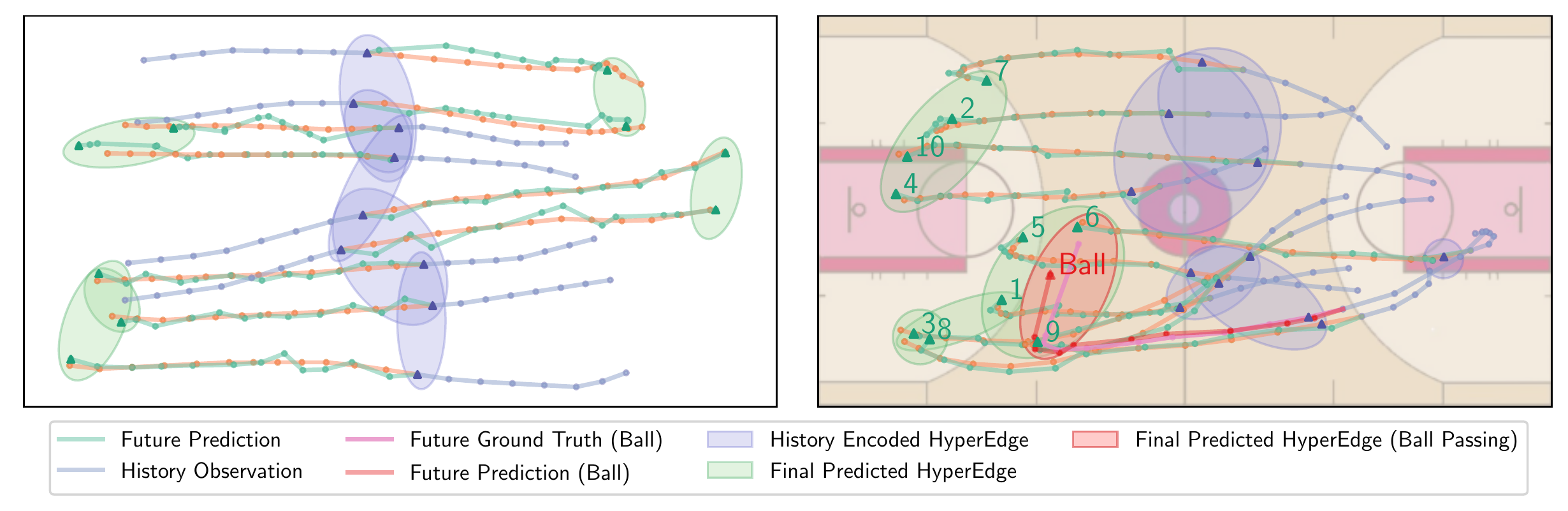}
	\vspace{-0.4cm}
	\caption{The visualization of results on the ETH / UCY dataset (left) and NBA dataset (right). The purple ellipses represent the hyperedges inferred based on historical observations while the green ones represent the hyperedges inferred at the final prediction step after evolution.}
	\vspace{-0.2cm}
	\label{fig:trajectory_plots}
\end{figure}

\vspace{-0.1cm}
\subsection{Qualitative analysis}
\vspace{-0.1cm}
We visualize the predicted trajectories and the group memberships inferred in the form of hyperedges for the crowd simulation, ETH / UCY, and NBA datasets in Figure \ref{fig:crowd_simulation}(c)(d) and Figure \ref{fig:trajectory_plots}, respectively.
For the crowd simulation, our approach can infer accurate hyperedges: The agents are predicted to split into two groups when getting close to the obstacle and our approach successfully infers the group membership for each agent. The significant improvement of inferred incidence matrices of hypergraphs, comparing second to first row in Figure \ref{fig:crowd_simulation}(c), can be related to the hypergraph evolution during prediction as well as the proposed regularizations on the inferred relations. 
Figure \ref{fig:trajectory_plots} shows the predicted trajectories and hyperedges on the ETH / UCY and NBA datasets, where our approach generates accurate trajectory hypotheses in both scenarios. 
The model can capture both group behaviors with similar motion patterns for pedestrians (left) and those with highly complex interactions for sports players (right). For example, in the NBA scenario, the predicted hyperedge at the final step, containing agents 6, 9, and the basketball, shows that our evolving hypergraph captures the relation of ball passing between teammates instead of only the spatial relations. 
We conclude that our approach infers reasonable and explainable evolution of groups.

\vspace{-0.1cm}
\section{Conclusion}
\vspace{-0.1cm}
In this paper, we present a group-aware dynamic relational reasoning approach and demonstrate its effectiveness for multi-agent trajectory prediction.
In addition to inferring the conventional graphs that models pair-wise relations between interacting agents, we propose to infer relation hypergraphs to model group-wise relations that exist widely in real-world scenarios such as human crowds or team-based sports. 
Different hyperedge types represent distinct group-wise relations. 
We employ a dynamic reasoning mechanism to capture the evolution of underlying relational structures.
Moreover, we propose effective mechanisms to regularize the smoothness of relation evolution over time and the sparsity of learned graphs / hypergraphs, which not only enhances training stability and robustness but also reduces prediction error.
The proposed approach infers reasonable relations and achieves state-of-the-art performance on both synthetic crowd simulations and multiple benchmark datasets.
A limitation of our method is that we need to preset the maximum number of hyperedges and the inferred group-wise relations could be missing or redundant. In future work, we will investigate how to learn flexible hyperedges without presetting constraints.
Also, our current approach only considers the agent-agent relations, we plan to further incorporate the environmental context information (e.g., map, road geometry) to generalize our approach to the scenarios where the traversable areas of interacting agents are constrained by obstacles or rules such as autonomous driving in the future.
Our work does not have negative societal impacts directly, but a rigorous test is suggested before it is applied to real-world tasks, especially in safety-critical situations where intelligent agents interact with humans intensively. 

\printbibliography

\end{document}